\pdfoutput=1

\documentclass[11pt]{article}

\usepackage{acl}

\usepackage{times}
\usepackage{latexsym}

\usepackage[T1]{fontenc}

\usepackage[utf8]{inputenc}
\usepackage{graphicx}
\usepackage{microtype}

\usepackage{inconsolata}

\usepackage{amsmath}
\usepackage{amsfonts}
\usepackage{algorithm}
\usepackage{algpseudocode}
\usepackage{booktabs}
\usepackage{multirow}
\usepackage{graphicx}
\usepackage{caption}
\usepackage{todonotes}
\usepackage[skip=0cm,list=true]{subcaption}

\title{Continuous Predictive Modeling of Clinical Notes and ICD Codes \\in Patient Health Records}

\author{Mireia Hernandez Caralt \And Clarence Boon Liang Ng \and Marek Rei \\
         Imperial College London, United Kingdom\\
         \normalfont{\texttt{\{mireia.hernandez-caralt22,clarence.ng21,marek.rei\}@imperial.ac.uk}}}

\author{
    {\bf Mireia Hernandez Caralt} ~~~~
    {\bf Clarence Boon Liang Ng} ~~~~
    {\bf Marek Rei} \\
    Imperial College London, United Kingdom \\
    \texttt{\small \{mireia.hernandez-caralt22,clarence.ng21,marek.rei\}@imperial.ac.uk}}

\begin{document}
\maketitle
\begin{abstract}

Electronic Health Records (EHR) serve as a valuable source of patient information, offering insights into medical histories, treatments, and outcomes. Previous research has developed systems for detecting applicable ICD codes that should be assigned while writing a given EHR document, mainly focusing on discharge summaries written at the end of a hospital stay. In this work, we investigate the potential of predicting these codes for the whole patient stay at different time points during their stay, even before they are officially assigned by clinicians. 
The development of methods to predict diagnoses and treatments earlier in advance could open opportunities for predictive medicine, such as identifying disease risks sooner, suggesting treatments, and optimizing resource allocation.
Our experiments show that predictions regarding final ICD codes can be made already two days after admission and we propose a custom model that improves performance on this early prediction task.

\end{abstract}

\section{Introduction}

Electronic health records (EHR) are rich repositories of patient information, chronicling their medical history, diagnoses, treatment plans, medications and outcomes \cite{jensen2012mining,johnson-etal-2016-mimic}.
The aggregation and modeling of this data over time presents a unique opportunity for revealing patterns that can improve patient care, operational efficiency, and healthcare delivery. 
Contained within the EHR are textual notes written by clinicians during patient encounters, which are essential for a comprehensive understanding of patient health. 
These free-text narratives stand out as a particularly rich source of nuanced information, but their unstructured format and domain-specific language use have left them largely underutilized, compared to more readily available structured data sources \cite{tayefi2021challenges}.


\begin{table}[]
\begin{tabular}{l|r|r}
\toprule
\textbf{Category} & \textbf{\# notes} & \textbf{\# words} \\
\midrule
Discharge summ. & 59,652                    & 79,649,691 \\ 
ECG               & 209,051                   & 5,625,393 \\ 
Echo              & 45,794                    & 12,810,062 \\ 
Nursing           & 1,046,053                                       & 185,856,841 \\ 
Physician         & 141,624                                        & 322,961,183 \\ 
Radiology         & 522,279                                        & 165,805,982 \\ 
Respiratory       & 31,739                                         & 11,957,187 \\ 
Other            & 26,988                                         & 13,086,023 \\ \bottomrule
\end{tabular}
\caption{The number of clinical notes from different categories, along with the number of words in those notes, in the MIMIC-III dataset.
}
\label{tab:notecateg}
\end{table}

Health records are often accompanied by the International Classification of Diseases (ICD) codes -- standardized codes that categorize diagnoses and procedures performed during clinical encounters \cite{cartwright2013icd}.
Assigning ICD codes manually is a highly time-consuming task necessary for billing, therefore previous research has been developing multi-label classification systems to detect applicable codes that should be assigned while writing a given document \cite{mullenbach-etal-2018-explainable,liu-etal-2021-effective}.
The research focus has been on classifying discharge summaries, which are written at the end of a patient's hospital stay \cite{ji-etal-2021-does,dai-etal-2022-revisiting}.
While this setup provides a useful proxy task, the complete EHR sequence is much longer, containing detailed reports from nursing and radiology, along with specialized notes on echographies and cardiograms (FCG), among others (Table \ref{tab:notecateg}). Recent work has argued that for most practical applications such code classification should be performed on earlier medical notes instead of the discharge summary \cite{cheng-etal-2023-mdace}.

In this work, we investigate the potential of predicting ICD codes for the whole patient stay at different time points during their stay. 
Beyond the task of detecting codes for a given note, we treat ICD codes as a structured summary of all the treatments provided and diagnoses assigned during a hospital stay.
The development of models for predicting this information early in the clinical timeline, based on partial indicators even before these codes have been officially assigned, would open many possibilities for predictive medicine.
Such systems would go beyond the post-discharge diagnostic practices and could be used for identifying early disease risks, suggesting potential treatments or optimizing hospital resource allocation.

We investigate the feasibility of this novel task and evaluate to what extent the final set of ICD codes can be predicted at earlier stages during the hospital stay.
In addition, we propose a custom model for this task that is able to improve prediction accuracy at different time steps.
Unlike previous ICD code prediction models, the architecture is designed with causal attention to ensure that representations at any point throughout a patient’s hospital stay are constructed based on the notes available up to that point, without accessing information in the future.
The model is then optimized to predict ICD codes after every additional note in the input sequence, instead of only at the discharge summary, teaching it to make predictions at any chosen time point during the hospital stay.
This task poses additional challenges, as the length of the complete EHR sequence far exceeds that of the discharge summary and early notes have a weaker correlation with the final labels. 
We introduce a novel method that both augments the data during training and extends the context during inference, 
substantially enhancing the performance on early ICD code prediction. The code for the model and the experiments are available online.\footnote{https://github.com/mireiahernandez/icd-continuous-prediction}

\section{Related Work}
The closest previous research to ours has been on automating ICD code assignment. Given a document mentioning diagnoses and treatments in free-form text, the aim is to detect the correct codes that should be assigned by the clinician. 
The first attempts at this task primarily relied on convolutional neural networks (CNNs) \cite{mullenbach-etal-2018-explainable, li-etal-2020-icd, liu-etal-2021-effective} and long short-term memory networks (LSTMs) \cite{RefWorks:RefID:8-vu2020label, yuan-etal-2022-code}. These models utilized pre-trained word2vec embeddings \cite{mikolov-etal-2013-efficient} and combined neighbouring word representations using convolutional filters or recurrent architectures. Despite their simplicity, some of these models achieved very high-performance baselines that were difficult to surpass with transformer approaches \cite{ji-etal-2021-does}.

Efforts to apply pre-trained transformers without further modifications to the ICD coding problem were unsuccessful \cite{ji-etal-2021-does,dai-etal-2022-revisiting}. The discharge summary contains 3,594 tokens on average, while the combined set of notes contains an average of 21,916 tokens per patient stay \cite{ng-etal-2023-modelling}. Crucial information to predict patient diagnoses is likely to be dispersed throughout these notes, thus models with limited context length risk overlooking a significant portion of relevant data. For this reason, subsequent studies focused on adapting transformer architectures to process longer textual sequences.

PubMedBERT-hier \cite{ji-etal-2021-does} employed hierarchical transformers to mitigate the length limitation issue, obtaining substantially better results.
This approach segments the document into chunks of 512 tokens and employs a BERT-based model pre-trained on the biomedical domain to encode each segment \cite{RefWorks:RefID:37-gu2022domain-specific}. The segments are then combined using a hierarchical transformer running over the CLS-token embeddings. The TrLDC model \cite{dai-etal-2022-revisiting} further improved performance
by employing a RoBERTa-based model pre-trained from scratch on biomedical articles and clinical notes \cite{RefWorks:RefID:63-lewis2020pretrained}. 

The PAAT (Partition Attention) model \cite{yoo-etal-2022-an} was able to surpass LSTM-based models like MSMN \cite{yuan-etal-2022-code} on the task of identifying the top 50 labels.
PAAT combines the Clinical Long-Former and a bi-LSTM, employing partition-based label attention for improved performance. HiLAT (Hierarchical Label-Attention) \cite{RefWorks:RefID:29-liu2022hierarchical}
achieved strong results on the top 50 labels by utilizing
ClinicalPlusXLNet, which outperforms other transformers like RoBERTa variants, with the downside being that the training speed is four times slower due to its bidirectional context capturing \cite{RefWorks:RefID:29-liu2022hierarchical}.

The HTDS (Hierarchical Transformer for Document Sequences) model \cite{ng-etal-2023-modelling} integrated earlier notes into the input context when making decisions about the discharge summary. 
This model employs a RoBERTa base transformer and a separate transformer layer running over the individual token representations, not only the CLS embeddings. 
They found that the earlier notes were indeed useful as additional evidence at the end of the hospital stay and provided performance improvements when classifying discharge summaries.


All this prior work has trained systems to assign ICD codes at the end of the hospital stay, whereas we investigate models for making predictions at any point during the stay.
Furthermore, while previous work has focused on detecting explicit mentions of diagnoses and treatments in a given text, we investigate to what extent future labels can be inferred based on only earlier documents.


\section{Architecture}
\label{sec:architecture}

\begin{figure*}[h!]
\centering
\includegraphics[width = 0.95\hsize]{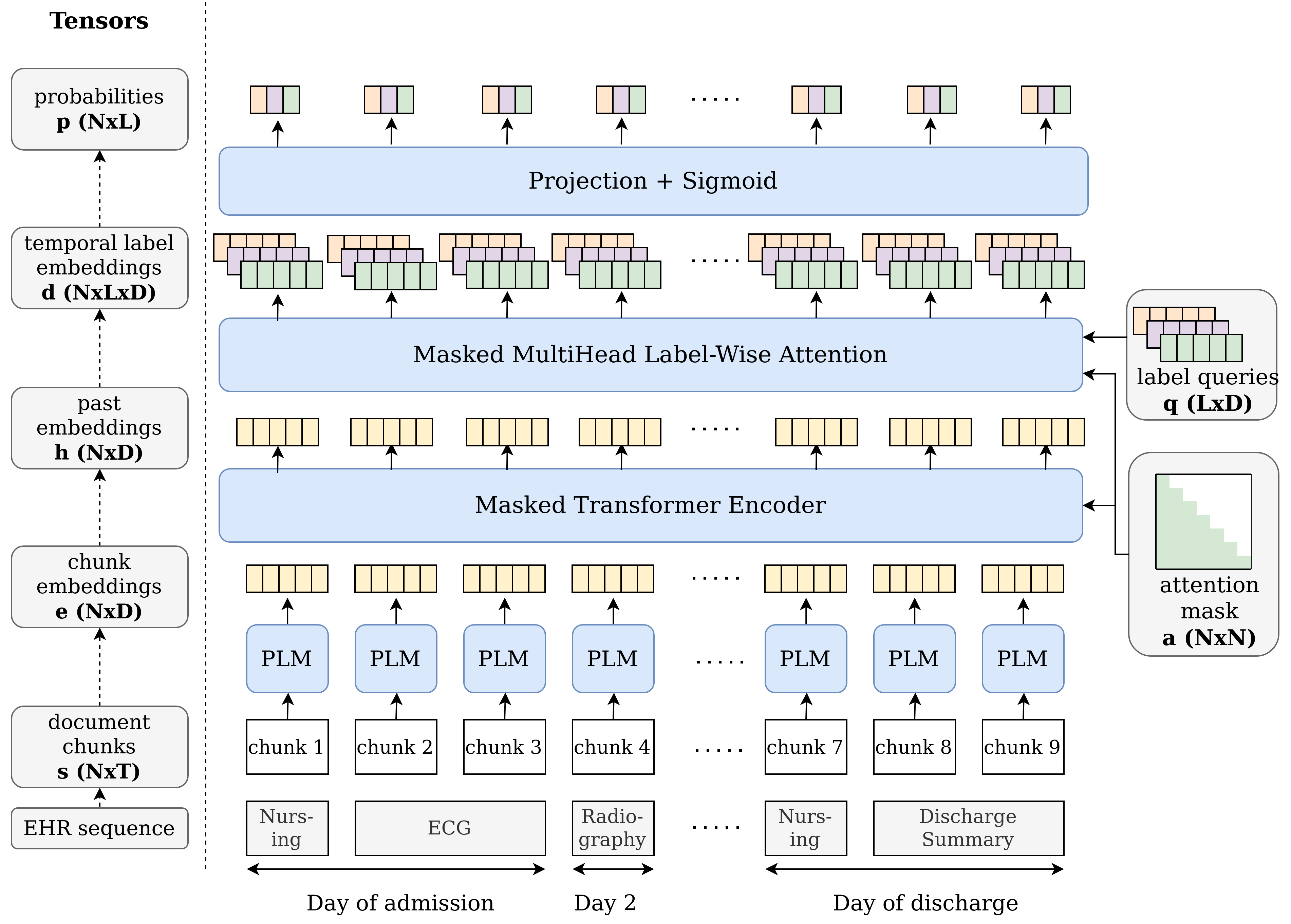}
\caption{\textbf{LAHST} (\textbf{L}abel-\textbf{A}ttentive \textbf{H}ierarchical \textbf{S}equence \textbf{T}ransformer) architecture. Clinical notes generated throughout the hospital stay are split into chunks. Each chunk is encoded using a pre-trained language model (PLM) to extract the CLS-token embedding. Next, a hierarchical transformer encoder is applied, utilizing causal masking to combine information among past segment embeddings. Finally, the network generates a distinct document representation for each label and temporal point combination and these are then transformed into probabilities by the output layer.}
\label{fig:model}
\end{figure*}

We investigate a model architecture that can be trained to encode a long temporal sequence of many clinical notes and make predictions at any time point, only using earlier notes as context. 
The model breaks the sequence into smaller chunks and encodes them using a hierarchical transformer. These chunks are combined with causal label attention, which gathers evidence with label-specific attention heads while ensuring that representations at any time are constructed based on the notes available up to that point, without accessing information in the future.
Finally, a probability distribution across the labels is predicted at each possible time point.
We refer to the model as a Label-Attentive Hierarchical Sequence Transformer (LAHST) and describe it in more detail below. Figure \ref{fig:model} provides a diagram of the architecture.



\textbf{Step 1. Document splitting}. Each document within the EHR sequence of a patient is tokenized and split into chunks of $T$ tokens. Each patient has a variable total number of chunks, and during training, a maximum of $N$ chunks is selected based on the criteria described in the next section.

\textbf{Step 2. Chunk encoding}.
Each of the chunks is encoded with a pre-trained language model (PLM), extracting the CLS-token embedding as the representation, yielding a tensor $e\in\mathbb{R}^{N\times D}$.
 We use the \texttt{RoBERTa-base-PM-M3-Voc} checkpoint, as it has been trained on two domains that match our task closely: 1) PubMed and PMC, which cover biomedical publications, and 2) MIMIC-III, which contains clinical health records \cite{RefWorks:RefID:63-lewis2020pretrained}.
 
\textbf{Step 3. Causal attention}. We augment a transformer layer with causal attention \cite{Choromanski2021FromBM} in order to combine temporal information from any previous step without providing access to information in the future steps. At the same time, the whole sequence can be efficiently processed in parallel by masking any attention connections on the right side of the target position.
This component takes as input the sequence of chunk embeddings \(e\in\mathbb{R}^{N\times D}\) and generates a sequence of embeddings $h\in\mathbb{R}^{N\times D}$ which combines information over past documents:
\begin{equation}
    h_i = CausalAttn(e_1, ..., e_i), i \in [1, N]
\end{equation}

\textbf{Step 4. Masked multi-head label attention.} We apply label-wise attention \cite{mullenbach-etal-2018-explainable} with two key modifications: the use of multiple attention heads, and the use of causal masking to obtain temporal label-wise document embeddings.
For each temporal position $t$, we define an attention mask $a_t \in \mathbb{R}^{L\times N}$ to prevent attention to future notes, which is constant in the label dimension, and nullifies attention weights beyond temporal position $t$.
\begin{equation}
a_t[:, i] = \begin{cases} 0, & \text{if } i \leq t \\ -\infty & \text{otherwise} \end{cases}
\end{equation}
 We then combine this mask with multi-head attention \cite{RefWorks:RefID:32-vaswaniattention} using learnable label embeddings $q\in\mathbb{R}^{L\times D}$ as queries and the previously generated past context embeddings $h\in\mathbb{R}^{N\times D}$ as keys and values:
\begin{equation}
\begin{split}
    d_t &= \text{MultiHeadAttn}(q, h, h, a_t) \\
    &=\text{Concat}(\text{head}_1, \cdots, \text{head}_H)W^o
\end{split}
\end{equation}

\noindent Here, each head inputs a linear projection of the key, query and value embeddings $e_{k,i}=W_i^K q$, $e_{q,i}=W_i^Q h$, $e_{v,i}=W_i^V h$ and applies masked attention \cite{Choromanski2021FromBM} as follows:
\begin{equation}
\begin{split}
    \text{head}_i &= \text{Attention}(e_{k,i}, e_{q,i}, e_{v,i}, \text{mask}=a_t) \\
    &=\text{SoftMax}(\frac{e_{k,i} e_{q,i}^T}{\sqrt{D/H}} + a_t)e_{v,i}
\end{split}
\end{equation}

This yields a sequence of label-wise document embeddings $d_t\in\mathbb{R}^{L\times\ D}$ for each position $t\in\{1,...,N\}$. In practice, we obtain all the embeddings $d\in\mathbb{R}^{N\times L\times D}$ efficiently in one pass by assigning the batch dimension to the temporal dimension.

\textbf{Step 5. Temporal label-wise predictions}.
Finally, temporal probabilities are calculated by projecting the embedding using linear weights $w\in\mathbb{R}^{L\times D}$ followed by a sigmoid activation. The probability at time $t$ for label $l$ is calculated using the the label weight $w_l\in\mathbb{R}^{D}$ and the label document embedding at position $t$, denoted as $d_{t,l}\in\mathbb{R}^D$:
\begin{equation}p_{t, l} = \text{sigmoid}( w_l \cdot d_{t,l})\end{equation}
The output of the model is a probability matrix $p\in\mathbb{R}^{N\times L}$, containing probabilities for each label at each temporal point. The masking process within the transformer and label attention modules ensures that time $t$ probability calculations consider only past documents.
The model is trained using the binary cross-entropy loss.

\section{Extending the Context}
\label{sec:eca}

Hierarchical transformer architectures break long inputs into smaller components and reduce the number of long-distance attention operations, thereby keeping memory and computation requirements more manageable when processing very long sequences. This makes them well-suited for ICD code classification, as local context is more important for this task and hierarchical models have been shown to outperform long-context models in this setting \cite{dai-etal-2022-revisiting}.
However, even hierarchical models have difficulty with very long sequences, particularly during training.
The gradient must be backpropagated through each individual chunk encoding, which can easily cause memory issues when the models are large and the number of segments exceeds a maximum limit.

For this reason, we propose a novel solution for applying hierarchical transformers to very long document sequences, such as the sequences of notes in EHR. We refer to this method as the Extended Context Algorithm (ECA). It consists of the following modifications to the training and inference loops.

\textbf{Training} (Algorithm \ref{alg:train}). For each episode of training, the loop iterates over the training dataset $\mathcal{D}_{train}$, processing each data sample $(s, y)$, where $s$ is the input sequence and $y$ is the corresponding label. Within the loop, a random selection of notes is chosen to create a subset of the input sequence, with the maximum number of chunks set as $N_{max}$. These sub-sequences $s'$ are then used for optimizing the model, each time sampling a slightly different training instance. Instead of trying to fit the whole sequence into the input during training, we sample notes and form multiple different shorter versions of the sequence for training the model. This has the added benefit of creating a data augmentation effect, as the model learns to make decisions based on different versions of the same datapoint.

\begin{algorithm}[t]
\caption{ECA Training loop}\label{alg:train}
\begin{algorithmic}
\State $\mathcal{D}_{train} \gets$ training set (sequence-label pairs)
\State $N_{max} \gets$ max. number of chunks
\For{each episode}
    \For{each $(s, y)$ in $\mathcal{D}_{train}$}
        \State $m \gets min(N_{max}, len(s))$ 
        \State select $m$ random indices ${i_1,...i_m}$
        \State sort ${i_1,\cdots,i_m}$ in ascending order
        \State $s' \gets [s[{i_1}],\cdots, s[{i_m}]]$
        \State $p, h$ $\gets \text{model.forward}(s')$ 
        \State $\mathcal{L}$ $\gets BCE(y, p)$
        \State do backward pass and optimizer step
    \EndFor
\EndFor
\end{algorithmic}
\end{algorithm}

\begin{algorithm}[t]
\caption{ECA Inference loop}\label{alg:inf}
\begin{algorithmic}
\State $\mathcal{D}_{test} \gets$ test set (no labels)
\State $N_{max} \gets$ max. number of chunks
\For{each $s$ in $\mathcal{D}_{test}$}
    \State $h_{list} \gets $empty list
    \For{$i$ in range$(0, len(s), N_{max})$}
        \State $s_{batch}\gets s[{i:i+N_{max}}]$
        \State $p_{batch}, h_{batch} \gets \text{model.forward}(s_{batch})$
        \State append $h_{batch}$ to $h_{list}$
        \EndFor
    \State $h \gets$ concatenate $h_{list}$ along batch dim.
    \State $p \gets$ model.label\_attention$(h)$
      
\EndFor
\end{algorithmic}
\label{alg:2}
\end{algorithm}

\textbf{Inference} (Algorithm \ref{alg:inf}). During inference, we process all the notes in the sequence in batches of $N_{max}$ chunks. Each sequence batch, denoted as $s_{batch}$, is encoded to obtain embeddings $h_{batch}\in\mathbb{R}^{N_{max}\times D}$. 
Even if the full sequence does not fit into memory, it can be processed in separate batches to obtain all the $h_{batch}$ embeddings.
These embeddings are then concatenated along the batch dimension to obtain chunk embeddings for the complete sequence $h\in\mathbb{R}^{N_{total}\times D}$. Finally, the collected embeddings are passed through causal attention and masked multi-head label attention to obtain predictions $p$ based on the complete sequence. 


As the computation can be performed in separate batches and then combined, this allows for considerably longer sequences to be used as input during inference. 
Unlike other methods for extending the context of transformers that rely on reducing or compressing long-distance attention \cite{beltagy-etal-longformer,munkhdalai2024leave}, this proposed method is also exact -- the result is always the same as it would be with a single pass using infinite memory.

\section{Experiment Set-up}

\subsection{Evaluation framework}

We investigate the novel task of temporal ICD code prediction, which requires the prediction of ICD codes at any point during the hospital stay using the notes available at that time, without relying on the discharge summary.
To evaluate the performance, we will compare the predictive power of our model at different points throughout the EHR sequence. Our evaluation setup is inspired by the ClinicalBERT model \cite{huang-etal-2019-clinicalbert}, which evaluates the likelihood of readmission at different cut-off times since admission.

The cut-off times were selected to be the 25\%, 50\%, and 75\% percentiles of the total volume of notes present in the training dataset, which are shown in Table \ref{tab:percentiles}. These correspond to 2, 5, and 13-day cut-offs, respectively. For example, in the \textit{2-day} setting, the model only has access to the notes written in the first 2 days in order to predict all the ICD codes that will be assigned to that patient by the end of their hospital stay. Where space allows, we additionally report on all the notes up to (but excluding) the discharge summary, indicating a setting where the model could be used to assist in the writing of the discharge summary itself. For comparison, we also report performance on the full sequence which includes the discharge summary, although this setting is retrospective and would not provide any predictive benefit.
In line with widely used approaches to ICD coding \cite{mullenbach-etal-2018-explainable}, we focus on Micro-F1, Micro-AUC and Precision@5 metrics, with additional metrics provided in the appendix.

\begin{table}[t]
\centering
\begin{tabular}{l|l|l}
\toprule
\textbf{Percentile} &\textbf{Days elapsed} & \textbf{\# notes} \\
\midrule
25\% & 1.8 & 112,594 \\
50\% & 5.2 & 225,160 \\
75\% & 12.8 & 337,726 \\
\bottomrule
\end{tabular}
\caption{Percentiles of the total volume of notes present in the training dataset. The number of days corresponding to the 25\textsuperscript{th}, 50\textsuperscript{th}, and 75\textsuperscript{th} percentiles will be used as temporal evaluation points throughout this project.}\label{tab:percentiles}
\end{table}

\subsection{Preprocessing}
We use the MIMIC-III dataset \cite{johnson-etal-2016-mimic} for evaluation, as it contains a collection of Electronic Health Records with timestamped free-text reports by nurses and doctors, together with the corresponding ICD-9 labels.
First, we follow the preprocessing steps outlined by the CAML approach \cite{mullenbach-etal-2018-explainable} to obtain a dataset of free-text clinical notes paired with ICD diagnoses and procedure codes, and we also extract their proposed train/dev/test splits. The label space is vast, so following their method, we focus on predicting the top 50 codes. 

For our novel task, we perform some additional preprocessing steps. First, we extract the timestamps of each note and, in cases where the specific time is missing, assign it to 12:00:00 of that day. Moreover, we found that some patients had additional notes beyond the discharge summary document, such as other discharge summaries or nursing notes. We exclude these additional notes to ensure that our EHR sequence always concludes with a single discharge summary document. 
We also exclude 14 patients as their EHR contains no other notes besides the discharge summary. 
Table \ref{tab:setup4} displays the statistics of our dataset at various temporal cut-offs. 

\begin{table}[t]
\centering
\begin{tabular}{l|c|c}
\toprule
\multicolumn{1}{c|}{} & \textbf{\# chunks / patient} & \textbf{\# patients} \\
\midrule
\textit{2 days} & 17.9\textsubscript{$\pm$22.1} & 1,559 \\
\textit{5 days} & 27.6\textsubscript{$\pm$33.9} & 1,559 \\
\textit{13 days} & 35.8\textsubscript{$\pm$42.4} & 1,559 \\
\textit{excl. DS} & 40.4\textsubscript{$\pm$47.7} & 1,559 \\
\textit{last day} & 48.4\textsubscript{$\pm$48.1} & 1,573 \\
\bottomrule
\end{tabular}
\caption{Length of EHR in number of chunks per patient (average and standard deviation) and count of patients of our dataset at different temporal cut-offs (dev set).}
\label{tab:setup4}
\end{table}


\subsection{Implementation details}

The model is implemented in Pytorch and it was trained on an Nvidia GeForce GTX Titan Xp (12GB RAM) GPU, utilizing an average memory of 11.22 GB. The model processed 5 samples per second and training took an average time of 11 hours and 50 minutes.  We used a super-convergence learning rate scheduler \cite{superconvergence}, based on its use in HTDS \cite{ng-etal-2023-modelling}, and an early-stopping strategy with a 3 epoch patience and a maximum of 20 epochs. Chunk size $T$ was set to 512 tokens as that is the largest size supported by \texttt{RoBERTa-base-PM-M3-Voc}. For the main experiments, a limit of $N_{max}=16$ was used during training, while the entire sequence (with up to $181$ chunks) was used for inference. The tuning ranges and chosen hyperparameter values are included in Appendix A.

\begin{table*}
\centering
\begin{tabular}{l|ccc|ccc|ccc|ccc}
    \toprule
      & \multicolumn{3}{c|}{\textbf{Last day}}                  & \multicolumn{3}{c|}{\textbf{0-13 days}}                  & \multicolumn{3}{c|}{\textbf{0-5 days}}             & \multicolumn{3}{c}{\textbf{0-2 days}}                   \\
      \midrule
      \textbf{Model} & \textbf{F1}            & \textbf{AUC}           & \textbf{P@5}           & \textbf{F1}            & \textbf{AUC}           & \textbf{P@5}           & \textbf{F1}            & \textbf{AUC}         & \textbf{P@5}         & \textbf{F1}            & \textbf{AUC}           & \textbf{P@5}            \\
      \midrule
TrLDC      & 70.1          & 93.7          & 65.9          & -             & -           & -           & -             & -             & -             & -           & -             & -             \\
PMB-H & 67.2          & 91.5          & 63.0          & 30.7          & 68.0        & 30.2        & 31.3          & 68.4          & 31.0          & 31.7        & 68.7          & 31.5          \\
HTDS       & \textbf{73.3} & \textbf{95.2} & \textbf{68.1} & 49.7          & 82.1        & 47.6        & 47.5          & 80.6          & 45.9          & 44.5        & 78.7          & 43.6          \\
HTDS*      & 70.7          & 93.8          & 66.2          & 48.6          & 82.0          & 47.0          & 46.7          & 80.7          & 45.5          & 43.6        & 78.7          & 43.3          \\
LAHST      & 70.3          & 94.6          & 67.5          & \textbf{52.9} & \textbf{87.0} & \textbf{53.0} & \textbf{50.3} & \textbf{85.4} & \textbf{50.7} & \textbf{46.0} & \textbf{82.9} & \textbf{47.1} \\
\bottomrule
\end{tabular}
\caption{Evaluation on the early ICD code prediction task at increasingly challenging temporal cut-offs. TrLDC \cite{dai-etal-2022-revisiting} result is from the respective paper. We evaluated PubMedBERT-Hier \cite[PMB-H; ][]{ji-etal-2021-does} and HTDS \cite{ng-etal-2023-modelling} at different early prediction points. HTDS* is a version of HTDS that is more comparable to LAHST in terms of computation requirements. LAHST is the model described in Sections \ref{sec:architecture} and \ref{sec:eca}. Results for PMB-H, HTDS, HTDS* and LAHST are averaged over 3 runs with different random seeds.}
\label{tab:mainresults}
\end{table*}

\section{Results}

In addition to the LAHST framework described in Sections \ref{sec:architecture} and \ref{sec:eca}, we also evaluate PubMedBert-Hier \cite{ji-etal-2021-does} and HTDS \cite{ng-etal-2023-modelling} on the early prediction task. HTDS was trained to consider earlier notes in the context while making decisions about the discharge summary, making it the most likely existing model to also perform well on the early prediction task. In addition, HTDS results are very close to the state-of-the-art on the MIMIC-III dataset, making it a very strong baseline. However, HTDS 
is a larger model and requires considerably more GPU resources compared to LAHST. Therefore, we also report a modified version (HTDS*) which has a comparable number of parameters.
We also include the performance of TrLDC \cite{dai-etal-2022-revisiting} from the respective paper as an additional strong baseline on classification of the discharge reports.

In Table \ref{tab:mainresults} we report the performance of these systems at increasingly challenging temporal cut-offs.
LAHST shows strong performance at any time point, outperforming all the other models at every early prediction task. 
The results indicate that some of the diagnosis and treatment codes for the whole hospital stay can be predicted already within the first few days of admission. While the performance of all systems is expectedly lower in the more challenging settings, they are still able to reach 46\% F1 and 82.9\% AUC with only 2 days of information, which could provide useful predictions to the hospital staff.
In all the early prediction settings, LAHST achieves the best results according to all metrics. While HTDS is trained to look at earlier documents and is also able to make competitive predictions, it is reliant on information in the discharge summary and therefore underperforms when this is not available. In contrast, the LAHST model is trained to make predictions based on varying amounts of evidence and achieves the best performance. 

In the "Last day" setting, which includes the discharge summary, HTDS slightly outperforms LAHST -- this is expected, as HTDS is a larger model and specifically trained for discharge summaries. However, when compared to the similarly-sized HTDS*, LAHST delivers comparable F1 along with improved AUC and P@5. Even though LAHST is not trained for this particular setting, the supervision on earlier time points helps it achieve good results also when classifying discharge summaries. In addition, it outperforms both PubMedBERT-Hier and TrLDC according to all metrics. 
We include larger results tables with additional metrics in Appendix B.



\section{Analyzing the Extended Context}

\textbf{Selection of context during inference}. The Extended Context Algorithm (ECA) allows the model to include much longer EHR sequences in the context during inference (with a generous 181 chunk cap applied in our experiments). We evaluate the effect of this algorithm compared to alternative strategies used in other hierarchical models.  The "Last" setting uses the most recent 16 chunks of text, illustrating the setting where the sequence is truncated from the beginning in order to fit into the model. The "Random" setting samples a random subset of  chunks from the sequence instead. The results in Table \ref{tab:ablateela} show that processing the entire sequence with ECA yields substantial performance improvements (+12.7, +13.3 and +15.4 Micro-F1 score for 2 days, 5 days and 13 days) compared to truncating or sampling the sequence. This result highlights the importance of including all the available notes in the input. Only when the discharge note is available (in the \textit{`Last day'} setting) the previous notes become less important and all the strategies give the same performance.

\begin{table}[t]
\centering
\begin{subtable}[c]{\linewidth}
\begin{tabular}{l|c|c|c}
\toprule
\textbf{} & \textbf{Last} & \textbf{Random} & \textbf{ECA} \\

\midrule
\textit{0-2 days}& 33.5 \textsubscript{$\pm$0.6} & 29.2 \textsubscript{$\pm$0.5}  & \textbf{46.2 }\textsubscript{$\pm$0.1}\\
\textit{0-5 days}& 37.6 \textsubscript{$\pm$0.3} & 35.1 \textsubscript{$\pm$0.4} &\textbf{50.9 }\textsubscript{$\pm$0.1} \\
\textit{0-13 days}& 38.2 \textsubscript{$\pm$0.1} & 37.7 \textsubscript{$\pm$0.5}&\textbf{53.6 }\textsubscript{$\pm$0.2}\\
\textit{Excl.DS} & 37.9 \textsubscript{$\pm$0.2} & 38.4 \textsubscript{$\pm$0.4} &\textbf{54.3 }\textsubscript{$\pm$0.2} \\
\textit{Last day}& 71.0 \textsubscript{$\pm$0.3} & \textbf{71.3}\textsubscript{$\pm$0.2} & 71.1 \textsubscript{$\pm$0.1} \\
\bottomrule
\end{tabular}
\vspace{1mm}

\end{subtable}
\caption{Micro-F1 score on the development set, using the LAHST model with alternative strategies for context inclusion.}
\label{tab:ablateela}
\end{table}

\textbf{Selection of context during training}. We investigate the effect of randomly sampling different sub-sequences of notes during training. We train an alternative version of LAHST by truncating the sequence to the most recent notes instead of sampling them randomly. During inference, both versions still receive all the notes as input, as described in Algorithm \ref{alg:inf}.
The results in Table \ref{tab:ablate} show how training without random sampling substantially decreases performance across all evaluation points (-8.4, -8.4, -8.2, -8.0, -0.9, F1 respectively). This indicates that randomly sampling different sub-sequences during optimization augments the training data with different variations which helps the model better generalize to different temporal cut-offs, without increasing memory or computation requirements.

\begin{table}[t]
\centering
\begin{tabular}{l|c|c}
\toprule
& \textbf{Last} & \textbf{Random/ECA} \\
\hline
\textit{0-2 days} & 37.8 \textsubscript{$\pm$0.2} & \textbf{46.2 }\textsubscript{$\pm$0.1} \\
\textit{0-5 days} & 42.5 \textsubscript{$\pm$0.1} & \textbf{50.9 }\textsubscript{$\pm$0.1} \\
\textit{0-13 days} & 45.4 \textsubscript{$\pm$0.1} & \textbf{53.6 }\textsubscript{$\pm$0.2} \\
\textit{Excl. DS} & 46.3 \textsubscript{$\pm$0.1} & \textbf{54.3 }\textsubscript{$\pm$0.2} \\
\textit{Last day} & 70.2 \textsubscript{$\pm$0.2} & \textbf{71.1 }\textsubscript{$\pm$0.1} \\
\bottomrule
\end{tabular}
\caption{Micro-F1 of LAHST on the development set, using alternative sampling strategies during training.}
\label{tab:ablate}
\end{table}

\section{Model Interpretability}
The attention weights in the label-attention layer of LAHST can potentially be used as an importance indicator of different input notes.
A higher weight is associated with an increased relevance of the particular document to predicting a specific code.
For an initial visualization, we average the weights across all the codes to find which document types are most important at different temporal cut-offs.

\begin{figure}[t]
\centering
\begin{subfigure}{.49\textwidth}
   \includegraphics[width=\textwidth]{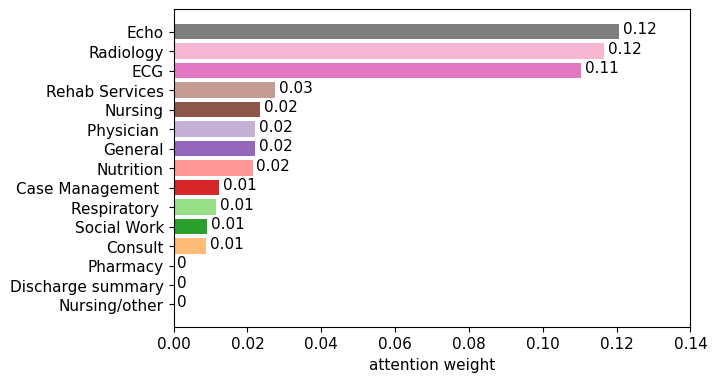}
   \caption{2-days cut-off}
\label{fig:g1}
\end{subfigure}
\hfill
\begin{subfigure}{.49\textwidth}
   \includegraphics[width=\textwidth]{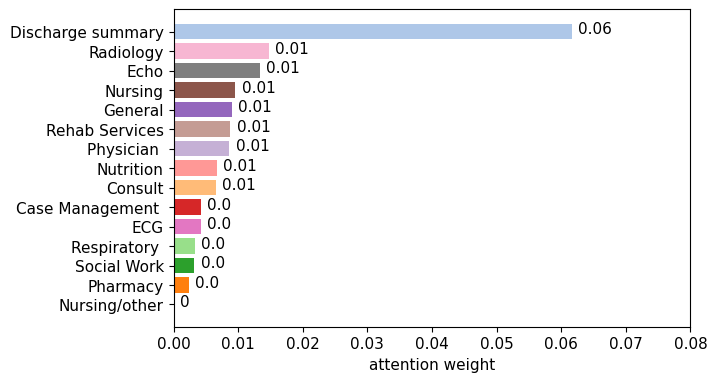}
   \caption{Last day cut-off}
\label{fig:g2}
\end{subfigure}
\caption{Average attention weight per document type at different temporal cut-offs. The LAHST model processes the complete EHR sequence and focuses more on reports of diagnostic tests for early prediction, switching to the discharge summary when it is available.}\label{fig:weightntote}
\end{figure}

The results are shown in Fig. \ref{fig:weightntote}. Within the 2-day cut-off, all the reports that have diagnostic characteristics have received the highest attention weights. For example, the echocardiography report is the description of an ultrasound test to identify abnormalities in the heart structure and is used by cardiologists to diagnose heart diseases \cite{RefWorks:RefID:67-van2023echocardiography}. 
The radiology reports detail the results of imaging procedures such as X-rays and MRIs to diagnose diseases \cite{alarifi-etal-2023-understanding}. 
All of such reports are highly technical and are specifically created to assist physicians with diagnostic practices. In the absence of the discharge summary, they are the most valuable document types for making early predictions of ICD-9 codes and the network has correctly focused more attention on them. In the "Last day" setting, the discharge summary becomes available, containing an overview of the entire hospital stay, and the same model is able to switch most of its attention to it.

\section{Conclusions}




In this study, we investigated the potential of predicting ICD codes for the whole patient stay at different time points during their stay. 
Being able to predict likely diagnoses and treatments in advance would have important applications for predictive medicine, by enabling early diagnosis, suggestions for treatments, and optimization of resource allocation.
We designed a specialized architecture (LAHST) for this task, which uses a hierarchical structure combined with label attention and causal attention to efficiently make predictions at any possible time points in the EHR sequence. 
The Extended Context Algorithm was further proposed to allow the model to better handle very long sequences of notes. 
The system is trained by sampling different sub-sequences of notes, which allows the model to fit into memory while also augmenting the data with variations of available examples.
During inference, the whole sequence is then processed separately in batches and combined together with a single attention layer, allowing for lossless representations of very long context to be calculated.

Our experiments showed that useful predictions regarding the final ICD codes for a patient can be made already soon after the hospital admission.  The LAHST model substantially outperformed existing approaches on the early prediction task, while also achieving competitive results on the standard task of assigning codes to discharge summaries. The model achieved 82.9\% AUC already 2 days after admission, indicating that it is able to rank and suggest relevant ICD codes based on limited information very early into a hospital stay.

\section{Limitations}
The primary focus of this project was to investigate the feasibility of this novel task and explore a novel architecture for the early prediction of ICD codes. Even though this could open up new avenues for early disease detection and procedure forecasting, our work has some limitations that should be considered in future work.

Firstly, our study is limited to the MIMIC dataset as it is one of the largest and most established available datasets containing electronic health records and ICD codes. However, the findings based on this dataset may not generalise equally to every clinical setting. Therefore, new experiments would need to be conducted on representative data samples before considering applying such technology in practice.

Our experiments focused on PubMedBERT-Hier \cite{ji-etal-2021-does}, HTDS \cite{ng-etal-2023-modelling} and LAHST. However, there are many other architectures and pre-trained models available which could be investigated in this setting.


Our model is based on a hierarchical transformer architecture which achieves good performance but is also quite computationally expensive compared to LSTM or CNN-based approaches (training the model took roughly 12 hours on a 12GB GPU). With our computational resources, we were limited to running experiments using the [CLS]-token representation and a maximum of 16 chunks in a batch.  However, with additional resources this work could be further scaled up by retaining all token representations and increasing the model size to allow for the allocation of additional chunks. 

Finally, our evaluation of the temporal ICD coding task is focused on reporting the aggregate metrics for the top 50 ICD-9 coding labels. Future work could investigate a larger number of labels, along with analysing the performance separately on individual labels and label types.

\section{Ethics Statement}
After careful consideration, we have determined that no ethical conflicts apply to this project. While clinical data is inherently sensitive, it is important to note that the MIMIC-III dataset has undergone a rigorous de-identification process, following the guidelines outlined by the Health Insurance Portability and Accountability Act (HIPAA). This de-identification process ensures that the dataset can be used for research purposes on an international scale \cite{johnson-etal-2016-mimic}.

While no conflicts were identified, machine learning systems for ICD coding carry certain risks when deployed in hospitals. Firstly, automated approaches are trained in a supervised manner using data from hospitals, and therefore, they are susceptible to reproducing manual coding errors. These errors may include miscoding due to misunderstandings of abbreviations and synonyms or overbilling due to unbundling errors \cite{aaron}. Moreover, automated systems may also suffer from distribution shifts, potentially affecting their portability across various EHR systems in different hospitals \cite{aaron}. To address these concerns, it is important to build interpretable models and develop tools that enable human coders to supervise the decisions made by ICD coding models.

\section*{Acknowledgements}
Mireia Hernandez Caralt acknowledges that the project that gave rise to these results received the support of a fellowship from ``la Caixa” Foundation (ID 100010434). The fellowship code is LCF/BQ/EU22/11930076.

\bibliography{anthology,custom}

\clearpage

\appendix

\section*{Appendix A}
\label{app:hyperparameters}

\begin{table}[h]
\begin{tabular}{l|c}
\hline
\textbf{Hyper-parameter}  & \textbf{Range} \\
\hline
Num. Layers (Mask. Transf.) & \textbf{1},2,3 \\
Num. Heads (Mask. Transf.) & \textbf{1},2,3 \\
Num. Heads (Label Atten.) & \textbf{1},2,3 \\
Peak LR & 1e-5, \textbf{5e-5}, 1e-4\\
\hline
\end{tabular}
\caption{The range of hyperparameters searched for tuning the model. The chosen value is shown in bold.}
\label{tab:hyper}
\end{table}

\section*{Appendix B}
\label{app:detailedresults}
Detailed results tables using different time cut-offs.

\begin{table}[h]
\centering
\begin{tabular}{l|ccccc}
\toprule
                                  & \multicolumn{5}{c}{0-2 days}                                                            \\
\midrule
                                  & Micro-F1 & Macro-F1 & Micro-AUC & Macro-AUC & P@5   \\
\midrule
PubMedBERT-Hier \cite{ji-etal-2021-does} & -                 & -                 & -                  & -                  & -              \\
TrLDC \cite{dai-etal-2022-revisiting}          & -                 & -                 & -                  & -                  & -              \\
HTDS \cite{ng-etal-2023-modelling}                       & 44.5              & 39.7              & 78.7               & 77.5               & 43.6           \\
HTDS*                             & 43.6              & 40.0              & 78.7               & 76.1               & 43.3           \\
LAHST                             & \textbf{46.0}     & \textbf{40.1}     & \textbf{82.9}      & \textbf{79.5}      & \textbf{47.1} \\
\bottomrule
\end{tabular}
\end{table}

\begin{table}[h]
\centering
\begin{tabular}{l|ccccc}
\toprule
                                  & \multicolumn{5}{c}{0-5 days}                                                            \\
                                  \midrule
                                  & Micro-F1 & Macro-F1 & Micro-AUC & Macro-AUC & P@5   \\
                                  \midrule
PubMedBERT-Hier \cite{ji-etal-2021-does} & -                 & -                 & -                  & -                  & -              \\
TrLDC \cite{dai-etal-2022-revisiting}          & -                 & -                 & -                  & -                  & -              \\
HTDS \cite{ng-etal-2023-modelling}                & 47.5              & 42.5              & 80.6               & 79.5               & 45.9           \\
HTDS*                             & 46.7              & 42.5              & 80.7               & 78.2               & 45.5           \\
LAHST                             & \textbf{50.3}     & \textbf{44.6}     & \textbf{85.4}      & \textbf{82.2}      & \textbf{50.7} \\
\bottomrule
\end{tabular}
\end{table}

\clearpage

\begin{table}[h]
\centering
\begin{tabular}{l|ccccc}
\toprule
                                  & \multicolumn{5}{c}{0-13 days}                                              \\
                                  \midrule
                                  & Micro-F1 & Macro-F1 & Micro-AUC & Macro-AUC & P@5   \\
                                  \midrule
PubMedBERT-Hier \cite{ji-etal-2021-does} & -             & -             & -           & -             & -            \\
TrLDC \cite{dai-etal-2022-revisiting}          & -             & -             & -           & -             & -            \\
HTDS \cite{ng-etal-2023-modelling}        & 49.7          & 44.6          & 82.1        & 81.2          & 47.6         \\
HTDS*                             & 48.6          & 44.6          & 82.0        & 79.7          & 47.0         \\
LAHST                             & \textbf{52.9} & \textbf{47.3} & \textbf{87.0} & \textbf{83.8} & \textbf{53.0} \\
\bottomrule
\end{tabular}
\end{table}

\begin{table}[h]
\centering
\begin{tabular}{l|ccccc}
\toprule
                                  & \multicolumn{5}{c}{Excl DS}                                                             \\
                                  \midrule
                                  & Micro-F1 & Macro-F1 & Micro-AUC & Macro-AUC & P@5   \\
                                  \midrule
PubMedBERT-Hier \cite{ji-etal-2021-does} & -                 & -                 & -                  & -                  & -              \\
TrLDC \cite{dai-etal-2022-revisiting}         & -                 & -                 & -                  & -                  & -              \\
HTDS \cite{ng-etal-2023-modelling}       & 50.2              & 45.2              & 82.3               & 81.5               & 47.8           \\
HTDS*                             & 49.0              & 45.1              & 82.4               & 80.2               & 47.5           \\
LAHST                             & \textbf{53.5}     & \textbf{47.8}     & \textbf{87.3}      & \textbf{84.1}      & \textbf{53.7} \\
\bottomrule
\end{tabular}
\end{table}

\begin{table}[h]
\centering
\begin{tabular}{l|ccccc}
\toprule
                                  & \multicolumn{5}{c}{Last Day}                                                            \\
                                  \midrule
                                  & Micro-F1 & Macro-F1 & Micro-AUC & Macro-AUC & P@5   \\
                                  \midrule
PubMedBERT-Hier \cite{ji-etal-2021-does} & 68.1              & 63.3              & 90.8               & 88.6               & 64.4           \\
TrLDC \cite{dai-etal-2022-revisiting}          & 70.1              & 63.8              & 93.7               & 91.4               & 65.9           \\
HTDS \cite{ng-etal-2023-modelling}         & \textbf{73.3}     & \textbf{67.7}     & \textbf{95.2}      & \textbf{93.6}      & \textbf{68.1}  \\
HTDS*                             & 70.7              & 64.9              & 93.8               & 91.6               & 66.2           \\
LAHST                             & 70.3              & 64.3              & 94.6               & 92.6               & 67.5   \\
\bottomrule
\end{tabular}
\end{table}

\end{document}